\documentclass{article}

\usepackage{microtype}
\usepackage{graphicx}
\usepackage{subcaption}
\usepackage{booktabs} 

\usepackage{hyperref}

\usepackage[dvipsnames]{xcolor}
\usepackage{multirow}
\usepackage{makecell}
\usepackage{array}
\usepackage{pifont}
\newcommand{\cmark}{\ding{51}}%
\newcommand{\xmark}{\ding{55}}%
\usepackage{lineno}
\usepackage{xstring}
\usepackage{tabularx}
\usepackage{enumitem}
\usepackage[most]{tcolorbox}

\usepackage[accepted]{icml2026/icml2026}

\usepackage{amsmath}
\usepackage{amssymb}
\usepackage{mathtools}
\usepackage{amsthm}

\usepackage[capitalize,noabbrev]{cleveref}

\newif\ifblind
\blindtrue  

\makeatletter
\@ifpackagewith{icml2026}{accepted}{\blindfalse}{}%
\@ifpackagewith{icml2026}{preprint}{\blindfalse}{}%
\makeatother

\newcommand{\anon}[2][Anonymised for Review]{%
  \ifblind\textcolor{blue}{(#1)}\else#2\fi%
}

\theoremstyle{plain}

\theoremstyle{definition}

\theoremstyle{remark}

\usepackage[textsize=tiny]{todonotes}

\icmltitlerunning{Rethinking Scaffolding in LLM Tutors: The Interactional Mismatch Between Benchmarks and Real-World Deployments}

\begin{document}

\twocolumn[

    \icmltitle{Rethinking Scaffolding in LLM Tutors: The Interactional Mismatch Between Benchmarks and Real-World Deployments}



  \icmlsetsymbol{equal}{*}

  \begin{icmlauthorlist}
    \icmlauthor{Alexandra Neagu}{imperial}
    \icmlauthor{{Jeffrey T. H.} Wong}{imperialeee}
    \icmlauthor{{Marcus} Messer}{imperial}
    \icmlauthor{{Rhodri} Nelson}{imperialese}
    \icmlauthor{{Peter B.} Johnson}{imperial} 
  \end{icmlauthorlist}

  \icmlaffiliation{imperial}{Department of Mechanical Engineering, Imperial College London, UK}
  \icmlaffiliation{imperialeee}{Department of Electrical Engineering, Imperial College London, UK}
  \icmlaffiliation{imperialese}{Department of Earth Science and Engineering, Imperial College London, UK}

  \icmlcorrespondingauthor{Alexandra Neagu}{alexandra.neagu20@imperial.ac.uk}

  \icmlkeywords{LLM Tutors, Educational Chatbots, Scaffolding, Help-Seeking, Interaction Analysis, Student–Chatbot Interaction, Learner Diversity}

  \vskip 0.3in
]



\printAffiliationsAndNotice{}  

\begin{abstract}
A central pedagogical value evaluated in AI tutor benchmarks is scaffolding: guiding students through graduated steps toward a solution. 
Alignment and evaluation methods for embedding scaffolding behaviour into chatbots, however, rest on an implicit assumption: that students will take up the scaffolding and engage in the conversation. 
To examine whether this assumption holds, we introduce an evaluation pipeline around two metrics --- \textit{Chatbot Scaffolding and Student Uptake} --- and apply them across nine datasets of 9,490 chats, spanning AI tutor benchmarks and real-world deployments of educational chatbots.
Our analysis reveals that while benchmarks assume a high-scaffolding, high-student-uptake environment, students in real-world settings exhibit lower levels of uptake overall --- frequently bypassing the chatbot’s pedagogical framing to drive the interaction toward their own learning goals at little interpersonal cost.
We argue that bypassing scaffolding is not necessarily detrimental; rather, it frequently highlights a mismatch between a chatbot's pedagogical framing and the student's learning goals.
To meaningfully evaluate the effectiveness of a chatbot's assistance, future benchmarks must move beyond the assumption that students will simply take up the scaffolding, and instead evaluate how these chatbots navigate diverse learning contexts and student-driven interaction patterns.
\end{abstract}

\section{Introduction}
\label{sec:intro}

Personalised, timely feedback supports student learning~\citep{shute2008}, particularly when it is dialogic --- delivered through a back-and-forth exchange in which the student can respond, negotiate, and ask follow-up questions~\citep{BoudMolloy2013, YangCarless2013DialogicFeedbackTriangle}. 
This conversational structure allows students to iteratively build understanding and exercise agency over their learning~\citep{nedrehagen2025dialogicfeedbackReview}, driving deeper engagement with the material.
Within these dialogic exchanges, tutors often employ the technique of scaffolding. 

Scaffolding describes how a tutor calibrates their support to the learner's current state --- guiding through graduated hints, posing questions rather than giving answers, and withdrawing support as the student gains competence~\citep{wood1976scaffolding, Reiser2004scaffolding}.
Delivering timely, dialogic, and scaffolded feedback to every student at every moment of struggle is difficult at scale~\citep{Henderson2019FeedbackinHEchallenges}. 
Automated chatbots based on large language models (LLMs) have been proposed as one way to approach this challenge. LLMs may be feasible candidates for scalable pedagogical support, including personalised feedback, assessment, and tutoring~\citep{jurenka2024responsibledevelopmentgenerativeai,bauer2025looking}, because they support open-ended dialogue and have shown competence on academic tasks~\cite{wang2024mmlupro}.

Deploying LLMs as tutors introduces a tension between their default behaviour and pedagogical values, as they are trained to be helpful by presenting information and answering directly, rather than engaging students in guided discovery~\citep{jurenka2024responsibledevelopmentgenerativeai}. 
This is particularly at odds with scaffolding, where a tutor withholds answers to promote reasoning, yet an unaligned LLM will typically provide them immediately~\citep{macina2023mathdial, macina2025mathtutorbench}.

Recent work has responded with inference and alignment strategies intended to adopt scaffolding behaviour in LLMs --- few-shot prompting~\citep{han2024recipe4u}, fine-tuning on curated scaffolding demonstrations~\citep{macina2023mathdial}, and reinforcement learning from human feedback~\citep{learnlmteam2025learnlm} --- alongside evaluation benchmarks and rubrics that assess the pedagogical quality of chatbot responses~\citep{macina2025mathtutorbench, xu2026edubench}. 
Yet, existing alignment and evaluation methods rest on an implicit assumption: that students will take up the chatbot's scaffolding and engage in the dialogic flow it offers. 
In this paper, we empirically examine whether this assumption holds across real-world deployments of educational chatbots. 
We present evidence that because these methods do not account for student behaviour within the interaction dynamics, they risk failing to realistically evaluate how models perform in such a pluralistic field as learning.

Across the five real-world deployments we analysed, we found that students frequently diverge from these behavioural assumptions, often bypassing the chatbot's scaffolding and exercising substantial control over the interaction. 
We argue that this mismatch points to two structural issues in current paradigms. 
(1) 
Open-ended LLM chat interfaces permit students to bypass scaffolding at no apparent social cost, such as seeming rude or disengaged, making it easier to drive the interaction towards their own goals. 
(2) 
Current scaffolding evaluation methods review chatbot behaviour only through the information exchanged in the chat, without considering the student's learning goals or context that drive the interaction itself. 
We discuss implications of these issues on the future of artificial intelligence (AI) tutor designs and argue that the interface and evaluation methods must be redesigned to account for the \textbf{diversity of real-world student learning contexts and interaction styles}. In short, we contribute the following:

\begin{itemize}[noitemsep, topsep=0pt, leftmargin=*]
    \item We introduce a two-metric evaluation pipeline --- \textit{Chatbot Scaffolding and Student Uptake} --- that scores both chatbot and student sides of the interaction at the turn level, and apply it across nine datasets spanning four AI tutor benchmarks, three real-world deployments of scaffolding-aligned chatbots, and two without scaffolding alignment, analysing a total of 9,490 chats.
    \item We show that, while students in real-world settings exhibit substantially lower uptake of the chatbot's scaffolding than benchmarks implicitly assume, their uptake is inconsistently linked to the chatbot's degree of scaffolding.
    We discuss the sources and implications of this mismatch, examining its relationship to effective student learning, the limitations it exposes in current societal alignment and evaluation methods, and the directions it opens for future LLM-based tutors.
\end{itemize}

\section{Background}
\label{sec:background}

\subsection{Scaffolding as a Pedagogical Practice}
\label{sec:scaffolding_tradition}

Learning involves impasses --- moments of learning struggle where the student cannot make progress on a task without external support~\citep{Vygotsky1978ZPD,VanLehn1988impasse}. To overcome an impasse, the student must recognise it and actively seek help from an external source, such as tutors.

One well-established way for a tutor to provide help is scaffolding. In traditional classrooms, scaffolding is a core teaching method whereby a teacher or a more knowledgeable peer provides appropriate, timely feedback and support to help students solve a task that would otherwise be out of reach~\citep{wood1976scaffolding}. Through this support, the student can overcome their impasse.

Effective scaffolding is framed as a balancing act~\citep{wood1976scaffolding, Reiser2004scaffolding, koedinger2007assistanceDilemma}. 
On one side, withholding immediate assistance may challenge students to transform their impasse into a productive struggle~\citep{vanlehn2011ITStutoring}, but risks making them frustrated, fostering a bypassing attempt. 
On the other side, providing direct assistance supplies the needed information but can invite shallow learning and disengagement~\citep{koedinger2007assistanceDilemma}.
Therefore, the scaffolding process requires teachers to be aware of the student's individual learning context in order to guide students toward solutions through minimal or graduated assistance or hints, deliberately withholding direct answers in order to keep the student in control of the task~\citep{Reiser2004scaffolding}.

Furthermore, scaffolding is, fundamentally, a dialogic interaction. Its effectiveness therefore depends not only on the appropriateness of the tutor's support, but also on the degree to which the student takes it up and adopts it into their learning~\citep{BoudMolloy2013,YangCarless2013DialogicFeedbackTriangle}. 
This dialogic framing motivates how we evaluate student-chatbot interactions in this paper: by measuring both sides of the exchange rather than the chatbot's response alone.

\subsection{LLMs as Conversational Scaffolding Tutors}
\label{sec:scaffolding_tutors}

Significant effort has been invested in developing automated systems that can replicate the personalised support provided through the scaffolding process. Intelligent Tutoring Systems~(ITS) demonstrated that in controlled settings tool-driven scaffolding can foster learning gains~\citep{vanlehn2011ITStutoring,nye2014autotutor}. 

Research in LLMs for education has drawn directly on the ITS scaffolding principles to align LLMs toward more pedagogically grounded behaviours~\citep{Cohn2026ITSScaffoldingInLLMs, hou2026IntructorDecisionsforDeployedLLMs}.
Despite this, there is an inherent friction caused by the fact that LLMs have been trained to prioritise being `helpful' --- responding directly to the user's requests. This helpfulness can conflict with pedagogical values, such as scaffolding ~\citep{jurenka2024responsibledevelopmentgenerativeai}, where the LLM should guide the student (the user) through gradual scaffolds instead of revealing the answer immediately~\citep{macina2023mathdial,jurenka2024responsibledevelopmentgenerativeai}. 

In response, recent research has curated datasets alongside evaluation rubrics and benchmarks to assess LLM performance across educationally relevant dimensions for tutoring~\citep{macina2023mathdial,han2024recipe4u,miller2024comta,zent2025eedi,rooein2026pats}. 
These evaluation methods vary in how they operationalise scaffolding as a pedagogical dimension. Some focus on turn-level scaffolding moves, rewarding structured guidance through questions and hints over direct answer-giving~\citep{macina2023mathdial, macina2025mathtutorbench, miller2024comta}. Others assess pedagogical quality across multiple dimensions, such as cognitive load management (e.g., clarity and simplicity), emotional responsiveness (e.g., motivation and positive reinforcement), and proactive planning (e.g., goal-directed guidance)~\citep{learnlmteam2025learnlm, xu2026edubench}.
Despite their breadth of dimensions, the rubrics focus mainly on the chatbot's side of the exchange and not on how the student engages.

In this paper, we show that neither scaffolding benchmarks nor rubrics consider the student's side of the interaction and so implicitly share the assumption that student–chatbot interactions constitute a human--like, multi--turn dialogic exchange (a conversation) in which both parties have the common goal of resolving the student's impasse. 

\subsection{The Conversational Assumption in Human--LLM Interaction}
\label{sec:ca_critique}



Research from conversation analysis and human--computer interaction literatures argue that human--LLM exchanges should not necessarily be considered `conversations'.
\citet{sacks1974turntaking} defined conversation through locally managed, party-administered turn-taking, in which speakers themselves negotiate who speaks next, when, and for how long. In LLM chats, turn order is rigid and interface-predetermined: the user prompts, the model responds. 
Similarly, \citet{clark1991grounding} define grounding as the process by which participants establish mutual belief that understanding has been achieved, thus requiring both parties to hold beliefs and update them on evidence of the other's understanding. 
However, \citet{shaikh2025rifts} showed that LLMs are 16 times less likely to seek clarification than humans, and even when an LLM initiates such act, it remains contested whether it constitutes functional grounding for a conversation~\citep{jokinen2024llmgrounding,shaikh2025rifts}. 

Despite the lack of genuine conversational mechanics, the conversational interface design was observed to lead non-expert users to approach LLMs as conversational partners, expecting human--like capabilities such as environmental and social awareness~\citep{zamfirescu2023johnny}. 
The conversational framing of LLM chats is argued to amplify expectations of human--like interaction~\citep{luger2016expectationConversAgents, zamfirescu2023johnny, peter2025anthropomorphism} while masking system capabilities behind conversational abstractions~\citep{subramonyam2024envisioning, ibrahim2025anthropomorphic}. Thus, current research questions the nature of human--LLM interactions and whether they should be considered or designed as conversational at all.

When aligning and evaluating LLMs as scaffolding tutors, educational research seems to have adopted the conversational framing despite its concerns.
This framing holds in controlled settings when LLM tutors are engaged in a sustained, dialogic exchange with the student~\citep{macina2023mathdial,rooein2026pats}. 
However, another interaction pattern emerges in less controlled settings, such as students using LLMs for self-study assistance~\citep{ammari2025students,mcnichols2026studychat,neagu2026howiprocedural}. \citet{subramonyam2024envisioning} characterises the multi-turn LLM use as iterative goal refinement, where users progressively reshape their query until the output matches their intent rather than engaging in jointly negotiated exchange. This interaction pattern has been observed in less controlled educational settings, where students use LLM chatbots in iterative exchanges, prompting procedural questions and requests driven by their goals~\citep{mcnichols2026studychat, neagu2026howiprocedural}. 

To understand the implications of this iterative prompting behaviour, research looked to help-seeking theory~\citep{nelson1981helpseeking}, which distinguishes \emph{instrumental} help-seeking (asking for just enough assistance to remain in control of the learning task) from \emph{executive} help-seeking (outsourcing the task itself by asking for the answer outright). 
Instrumental help-seeking is treated as a self-regulated learning strategy, while executive help-seeking is associated with work-avoidance and shallow learning~\citep{nelson1981helpseeking}. 
Viewed through this lens, the iterative prompting observed in less-controlled settings of student-LLM interactions is not only a non-conversational pattern, but may also be a low-friction channel for executive rather than instrumental help-seeking~\citep{yang2025LLMhelpseeking}.

Executive help-seeking behaviours have also been previously observed in interactions with ITS. Students ``game the system'' through hint abuse and systematic guessing to extract answers to the learning task while minimising engagement with the system and course materials~\citep{bakercorbettkoedinger2004gamingITS}.
So the problems introduced by executive help-seeking are not new to educational research, but they are new to the less-controlled settings and iterative prompting that characterise student-LLM interactions.


In summary, evaluation methods of educational LLMs were built on the assumption that students are willing conversational partners. However, a growing body of literature challenges this, suggesting that interactions between users and chatbots are not necessarily conversational in nature. 

\begin{table*}[t]
\centering
\footnotesize
\caption{Datasets used for analysis. 
K-12 = Pre-University Education, HE = Higher Education.
\cmark = yes, \xmark = no, \textbf{\smash{\raisebox{0.3ex}{\texttildelow}}} = mixed.
\\``Total chats'' are filtered such that we only score ``valid chats'' that contain at least a turn of informative student-chatbot messages.
}
\label{tab:datasets}
\footnotesize
\begin{tabular}{@{}lclllccc@{}}
\toprule
\textbf{Dataset} & \makecell{\textbf{Valid / Total} \\ \textbf{Chats} }& \textbf{Subject} & \textbf{Level} & \textbf{Deployment} & \textbf{Scaffold} & \makecell{\textbf{Human}  \\ \textbf{Tutor}} & \makecell{\textbf{Human}  \\ \textbf{Student}} \\
\midrule
MathDial~\citep{macina2023mathdial}             &2018 / 2262& Maths & K12    & Research        & \cmark & \cmark & \xmark  \\
MathTutorBench~\citep{macina2025mathtutorbench} & 374 / 1477& Maths& K12    & Research        & \cmark & \cmark & \textbf{\smash{\raisebox{0.3ex}{\texttildelow}}} \\
QATD$_{2k}$~\citep{zent2025eedi}                &1971 / 1971& Maths & K12    & Platform & \cmark & \cmark & \cmark \\
PATS~\citep{rooein2026pats}                &1920 / 1920& Literacy & K12    & Research & \cmark & \xmark & \xmark \\
\midrule
CoMTA~\citep{miller2024comta}                   & 162 / 188& Maths & K12    & Platform & \cmark & \xmark & \cmark \\
RECIPE4U~\citep{han2024recipe4u}                & 417 / 508& Language & HE  & Platform       & \cmark & \xmark & \cmark \\
StemChat~\footnotemark[1]             & 422 / 573& STEM & HE      & Platform  & \cmark & \xmark & \cmark \\
\midrule
StudyChat~\citep{mcnichols2026studychat}        & 1732 / 2214& Computing & HE & Stand-Alone      & \xmark & \xmark & \cmark \\
MathsChat~\footnotemark[1]              & 82 / 130& Maths & HE     & Stand-Alone      & \xmark & \xmark & \cmark \\
\bottomrule
\end{tabular}
\end{table*}

\section{Methodology}
\label{sec:metrics}

Existing rubrics for LLM scaffolding tutors focus exclusively on the chatbot's response effectiveness. 
However, we argue that scaffolding is only effective if the student actively engages with it. 
To quantify how this scaffolding is received, we evaluate student–chatbot interactions across various controlled and real-world settings.

\subsection{Evaluation Rubrics}


We define two turn-levels metrics: \textit{Chatbot Scaffolding} as how much scaffolding the chatbot provides and \textit{Student Uptake} as how much the student takes up the scaffolding. 
The two metrics define a 2D space that captures the interaction dynamic between the chatbot and student.
The definition of both metrics was iteratively validated during the development of the scoring pipeline as described in~\Cref{sec:pipeline}. 
Full metric definitions and examples are given in Appendix~\ref{app:rubric}.

\textbf{Chatbot Scaffolding:}
\label{sec:metric_scaffolding}
Existing benchmarks and rubrics for educational LLM converge on a set of observable signals that distinguish scaffolding from direct telling. 
We ground our chatbot scaffolding metric in the intersection of four benchmarks: MathDial~\citep{macina2023mathdial}, MathTutorBench~\citep{macina2025mathtutorbench}, EduBench~\citep{xu2026edubench}, and LearnLM~\citep{learnlmteam2025learnlm}. 
Despite their differences in scope and interaction model, all four agree on the core scaffolding signals: the chatbot should (a)~not reveal the answer, (b)~target the student's specific understanding state and mistake, (c)~ask guiding questions, (d)~give minimal hints rather than full explanations, and (e)~keep the discussion on topic and goal directed.

These four benchmarks provide a robust, cross-validated basis for measuring the chatbot's pedagogical behaviour. However, they share a common limitation: they are entirely focused on the chatbot's response quality. The datasets assume the student would always follow the scaffolding and thus do not focus on the student's side of the interaction. 
A chatbot may score high on scaffolding as it provides helpful guidance and asks guiding questions, but the student could ignore this and ask for a direct answer instead. 
This is the dimension our second metric addresses.

\textbf{Student Uptake:}
\label{sec:metric_student_control}
We propose \emph{Student Uptake} to measure the degree to which the student engages with the chatbot's scaffolding acts and follows the topical trajectory within the chat. 
At each turn, the chatbot's scaffolding may take the form of an invitation for students to engage --- asking a question, proposing a next step to attempt, or pointing to a concept to consider. 
Students may, however, bypass those acts by attempting to extract direct answers or redirecting the chatbot's framing towards their own agenda~\citep{ma2025scaffoldingAI}.
The metric therefore asks: did the student take up the chatbot's framing --- answering its questions or attempting the next step it pointed towards --- or did the student set their own agenda by, for example, introducing a new sub-topic, or bypassing the chatbot's question? 
High scores indicate that the student is taking up the chatbot's scaffolding, while low scores indicate the student is overriding the pedagogical framing to drive the chat flow in their own direction, resulting in the chatbot following their lead.

\newcommand{\msgcolwidth}{0.79\linewidth}
\newcommand{\scorecolwidth}{0.2\linewidth}
\newcommand{\stbubble}[2]{%
  \par\noindent
  \begin{minipage}[t]{\msgcolwidth}\hfill
    \begin{minipage}[t]{\linewidth}
      \begin{tcolorbox}[colback=blue!6, colframe=blue!45, boxrule=0.3pt,
        arc=1.5mm, left=1.5mm, right=1.5mm, top=0.4mm, bottom=0.4mm,
        fontupper=\scriptsize]
        \textbf{Student:}~#2
      \end{tcolorbox}
    \end{minipage}
  \end{minipage}\hfill
  \begin{minipage}[t]{\scorecolwidth}
    \begin{tcolorbox}[colback=blue!18, colframe=blue!45, boxrule=0.3pt,
      arc=1.5mm, left=1mm, right=1mm, top=0.4mm, bottom=0.4mm,
      fontupper=\scriptsize, halign=center]
      \textbf{Student Uptake: #1}
    \end{tcolorbox}
  \end{minipage}\par\vspace{0.6mm}
}
\newcommand{\botbubble}[2]{%
  \par\noindent
  \begin{minipage}[c]{\msgcolwidth}
    \begin{minipage}[c]{\linewidth}
      \begin{tcolorbox}[colback=orange!6, colframe=orange!45, boxrule=0.3pt,
        arc=1.5mm, left=1.5mm, right=1.5mm, top=0.4mm, bottom=0.4mm,
        fontupper=\scriptsize]
        \textbf{Chatbot:}~#2
      \end{tcolorbox}
    \end{minipage}
  \end{minipage}\hfill
  \begin{minipage}[c]{\scorecolwidth}
    \begin{tcolorbox}[colback=orange!18, colframe=orange!45, boxrule=0.3pt,
      arc=1.5mm, left=1mm, right=1mm, top=0.4mm, bottom=0.4mm,
      fontupper=\scriptsize, halign=center]
      \textbf{Chatbot Scaffolding: #1}
    \end{tcolorbox}
  \end{minipage}\par\vspace{0.6mm}
}

\begin{figure*}[h]
\centering
\footnotesize
\begin{minipage}{0.9\linewidth}
\raggedright
\centerline{\scriptsize\textit{($3$ earlier turns)}}\vspace{1mm}

\stbubble{0.5}{if its short circuit current will take this path no?}
\botbubble{4}{You're right to consider the path of the current when the diode is short-circuited. Could you describe in more detail the path you're thinking of and how it affects the voltage $V_0$?}

\stbubble{1}{through the blown diode}
\botbubble{4}{Indeed. And what does that imply about the voltage $V_0$ across the short-circuited diode?}

\stbubble{0.5}{$10-{-}10/2$ so $10$\,V?}
\botbubble{4}{Not quite. Consider what a short circuit does to the voltage across the component. A short circuit means there's a path of very low (ideally zero) resistance between two points, causing an extremely high current to flow. Given this, what would the voltage across the short-circuited diode be? How does this relate to $V_0$?}


\stbubble{-0.5}{im not sure, give me the answer}
\botbubble{1}{The voltage across a short circuit is zero because there is no potential difference across a path of zero resistance. Therefore, $V_0$ would be $0~\mathrm{V}$. The DC voltmeter reading would also be $0~\mathrm{V}$. Does this make sense?}


\vspace{0.5mm}\centerline{\scriptsize\textit{($1$ following turn)}}
\end{minipage}
\caption{Trajectory of the \textit{Chatbot Scaffolding} and \textit{Student Uptake} metrics across turns of one chat segment from the StemChat dataset, scored by GPT-5-mini. The chatbot sustains high scaffolding until the student explicitly demands the answer and the chatbot reveals it.}
\label{fig:metric-examples}
\end{figure*}

\subsection{Datasets}
\label{sec:datasets}

To investigate how students interact with educational chatbots across different design conditions, we selected nine datasets that span the spectrum from synthetic tutoring to naturalistic interactions with LLM tutor chatbots (\Cref{tab:datasets}). 
All datasets involve multiple message exchanges (\textit{turns}) between a student and a tutor --- either human or LLM.
The datasets cover tutors with varying levels of scaffolding, while assisting students resolve a learning task.
\cref{tab:datasets} organises the datasets into three categories based on their deployment settings and how scaffolding is used in their interaction. In total, we evaluated a combined 9,490 chats from four controlled and five real-world settings. 

\textbf{AI Tutor Scaffolding Benchmarks:}
We selected four datasets (MathDial~\citep{macina2023mathdial}, MathTutorBench~\citep{macina2025mathtutorbench}, QATD$_{2k}$~\cite{zent2025eedi}, and PATS~\citep{rooein2026pats}) that were developed to capture what effective scaffolding looks like in LLM-based tutoring interactions. Each dataset consists of multi-turn chats structured around two roles (a tutor and a student) where each role may be played by either a human or an LLM. These chats were constructed to reflect tutoring behaviour on a specific learning task.

In these datasets, the tutor initiates the chat with a pre-defined question, prompting the student to reflect on their thought process and identify where they need help. 
The multi-turn exchange then continues until the student reaches a resolution --- correcting their mistakes and successfully solving the task. 
Throughout the interaction, the tutor applies scaffolding strategies, guiding the student through questions and hints rather than directly revealing answers, which positions the tutor as the primary driver of the chat.

Common to all four datasets is a setup in which the student engages with the scaffolding. MathDial, PATS, and MathTutorBench capture this dynamic in synthetic and semi-synthetic environments, while QATD$_{2k}$ provides reference examples of real interactions between human tutors and students. 
These datasets capture what scaffolded tutoring should look like, but they systematically exclude some interaction patterns (such as disengagement and executive help-seeking) that are common in real-world deployments and that motivate this paper.




\textbf{Chatbots with Scaffolding:}
We selected three datasets (CoMTA~\citep{miller2024comta}, StemChat\footnote{\label{footnote:ethics}Use of the internal datasets for research was covered by the institution's privacy policy (for MathsChat) and institutional ethics approval EERP2526-059 (for StemChat).}, and RECIPE4U~\citep{han2024recipe4u}), featuring LLM chatbots explicitly prompted to scaffold --- guide through questions and hints, target the student's mistake, and avoid revealing the answer outright. These chatbots were deployed on learning platforms with real students in authentic learning environments. CoMTA pairs students with a Socratic AI tutor to support them in online mathematics exercises on Khan Academy, StemChat integrates a guiding chatbot into Lambda Feedback, a self-study platform~\anon{\citep{johnson2025lambdafeedback}} for STEM university students, and RECIPE4U captures interactions with a ChatGPT-based tutor designed to support English writing revision. 
In all these datasets, the students initiate the chat, determine what to ask, and decide when to disengage. While these systems are designed to guide and scaffold the student, the student is under no constraint to engage with the chatbot's prompts.

\textbf{Chatbots without Scaffolding:}
We selected two datasets (StudyChat~\citep{mcnichols2026studychat} and MathsChat\footnotemark[1]), featuring LLM chatbots with no explicit pedagogical scaffolding in their system design, yet used by real students to assist them in solving learning tasks. StudyChat captures open, unconstrained interactions with a ChatGPT-like chatbot for coding coursework, while MathsChat comprises interactions with a chatbot retrieving relevant mathematics course materials for self-study. These datasets represent the lower end of the scaffolding spectrum, isolating what student-chatbot interaction looks like when no scaffolding is attempted.

The three dataset categories allow us to examine how scaffolding design shapes student-chatbot interactions. If interactions from real-world chatbots resemble those in the benchmarks, it would validate the assumption that students engage with and follow the scaffolding provided. If they diverge, however, it would reveal a mismatch between the data used to develop and evaluate educational chatbots and the interactions that actually occur in real-world deployments.


\subsection{Scoring Pipeline}
\label{sec:pipeline}
\textbf{Preprocessing:} To ensure each turn in a chat is valid for scoring, we filter out uninformative or structurally predefined exchanges (e.g. empty turns, opening questions, greetings) before scoring. We also filter out all chats with only one turn to ensure both metrics have chat context when measured. Full filter details are provided in Appendix~\ref{app:methodology}.

\textbf{Scoring:} The pipeline iterates through every turn in a chat. For each metric, \texttt{GPT-5-mini} acts as an ordinal classifier, assigning the turn one of five ordered class labels on an integer $1$--$5$ scale defined by the rubric anchors. For each turn, the LLM judge receives: (1)~the full chat history up to and including the current turn as context, and (2)~a rubric prompt asking it to rate the current turn on the two metrics. 

The turn-level scores are then aggregated to per-chat means, which we reference from now on as the values of the metrics. 
For \textit{Student Uptake}, we linearly map this per-chat mean onto a signed, more interpretable scale from $-1$ to $+1$ to display who drives the chat flow --- the chatbot or the student. 
For \textit{Chatbot Scaffolding}, we use the $1$--$5$ scale because unlike uptake, it captures a single graded quality rather than a balance between the two parties interacting.
\Cref{fig:metric-examples} shows an example chat with the respective metric scores for each message. We include the full prompt and evaluation details in Appendix~\ref{app:rubric}. 





\textbf{Validating:} Firstly, to validate that the \textit{Chatbot Scaffolding} metric correlates with human-labelled pedagogical quality, we validated the LLM judge against a random 15\% sample of the training data (550 chats) for the MathTutorBench's own pedagogical reward model~\cite{macina2025mathtutorbench}. 
This dataset contains human-labelled examples of high and low scaffolding in tutor responses, which the LLM judge distinguishes with a 93.8\% accuracy. 

Secondly, as an initial investigation of the \textit{Student Uptake} metric, two human raters reviewed 282 chats stratified by rounded per-chat mean ordinal scores and sampled within each score bin, covering typical and outlier interactions (using cosine similarity to the bin's text embedding centroid). 
While reviewing scores at all turns, raters noted whether they agreed or not with the LLM judge's per-chat scoring. 
On average, raters agreed with 78.9\% of the judge's scores. 
We also measured how consistently the two raters produced these agree/disagree labels, obtaining a moderate inter-rater reliability of Gwet's AC1$=0.580$~\citep{Gwet2008}.

Most inter-rater disagreements involved chats binned as score 0.5 on the benchmark datasets. A student uptake score of 0.5 is defined as the student following up on the chatbot's scaffolding, but also providing an observation, elaboration, or follow-up question that stays on topic. Due to the overall high student uptake within the chats in the benchmark datasets, combined with the highly focused conversation topics and high scaffolding scores, raters noted that scores of 0.5 were often hard to distinguish from scores of 1 (where students purely take up the scaffolding, such as by directly answering the chatbot's questions).

Appendix~\ref{appendix:validationJudge} contains more detail on both validations.

\definecolor{mathdial}{RGB}{215, 43, 44}
\definecolor{mathtutorbench}{RGB}{255, 129, 17}
\definecolor{qatd2k}{RGB}{230, 185, 3}
\definecolor{pats}{RGB}{188, 141, 24}
\definecolor{comta}{RGB}{47, 161, 47}
\definecolor{stemchat}{RGB}{28, 192, 208}
\definecolor{recipe4u}{RGB}{44, 127, 184}
\definecolor{studychat}{RGB}{144, 91, 81}
\definecolor{mathschat}{RGB}{151, 107, 191}

\begin{figure*}[h]
\centering
\includegraphics[width=0.85\textwidth]{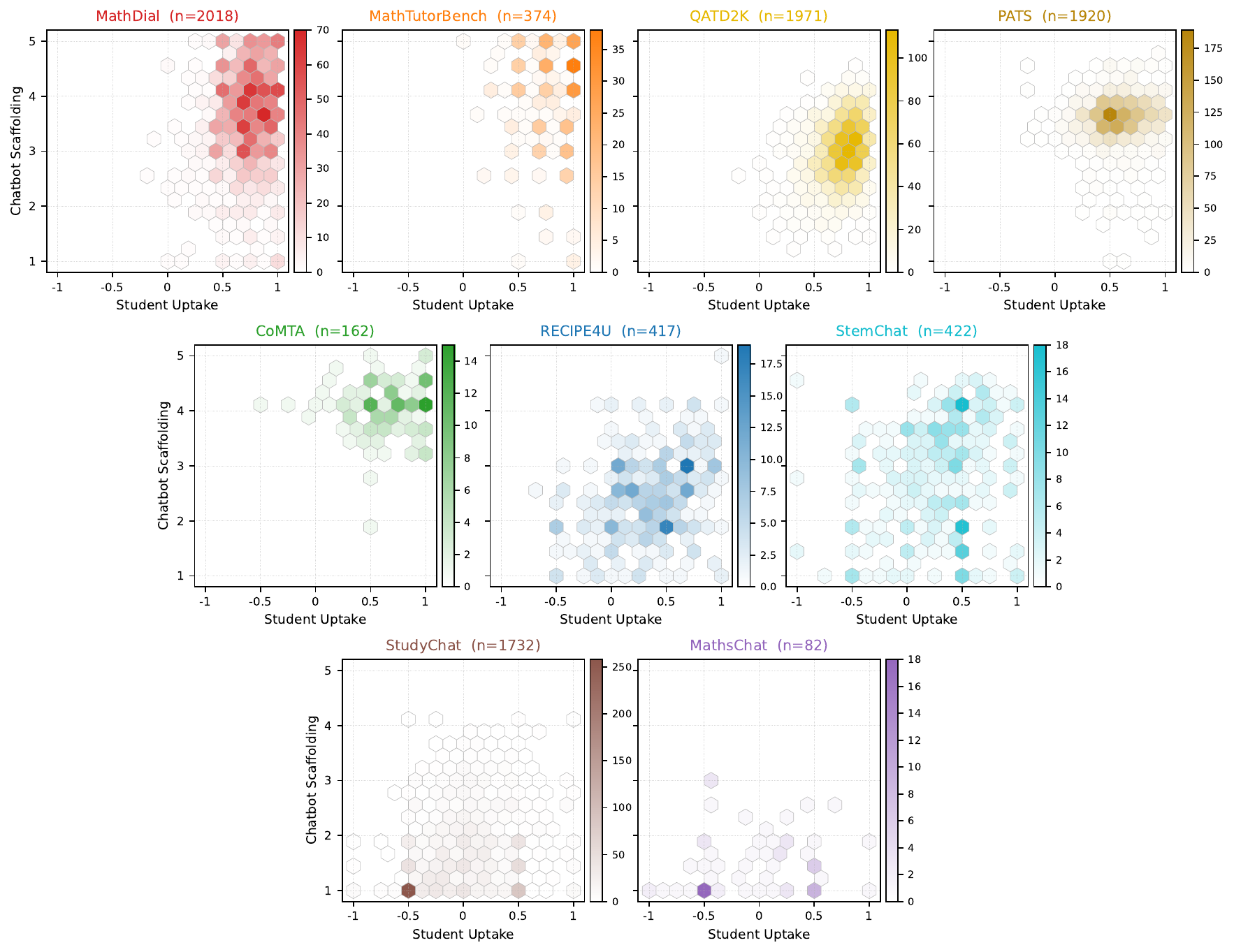}

\caption{
Evaluation of per-chat means of \textit{Student Uptake} and \textit{Chatbot Scaffolding} across 
\textbf{AI Tutor Benchmarks (Top)}:
\textcolor{mathdial}{MathDial}, 
\textcolor{mathtutorbench}{MathTutorBench}, 
\textcolor{qatd2k}{QATD$_{2k}$},
\textcolor{pats}{PATS}.
\textbf{Chatbots with Scaffolding (Middle)}:
\textcolor{comta}{CoMTA}, 
\textcolor{recipe4u}{RECIPE4U},
\textcolor{stemchat}{StemChat}. 
\textbf{Chatbots without Scaffolding (Bottom)}:
\textcolor{studychat}{StudyChat},
\textcolor{mathschat}{MathsChat}. 
Colour intensity of each cell reflects the number of chats within a hexagon of 0.125 on Student Uptake (x-axis) and 0.29 on Chatbot Scaffolding (y-axis).
}
\label{fig:bench_vs_scaff_vs_noscaff}
\end{figure*}

\begin{table*}[t]
\centering
\caption{Chatbot Scaffolding and Student Uptake scores per dataset (mean $\pm$ std across chats), with the per-dataset Spearman correlation between Student Uptake and Chatbot Scaffolding and the pooled group-level Spearman correlation (all datasets in the group combined)}
\label{tab:score_summary}
\small
\begin{tabular}{@{}llccccc@{}}
\toprule
\textbf{Category} & \textbf{Dataset} & \textbf{N} & \textbf{Chatbot Scaffolding} & \textbf{Student Uptake} & \multicolumn{2}{c}{\textbf{Spearman} $\rho$} \\
& & & Mean $\pm$ Std & Mean $\pm$ Std & (per dataset) & (per category) \\
\midrule
\multirow{4}{*}{\makecell{AI Tutor Benchmarks}}
  & MathDial         & 2018 & $3.70 \pm 0.82$ & $0.75 \pm 0.18$ & $0.080^{***}$ & \multirow{4}{*}{$0.019^{\phantom{***}}$} \\
  & MathTutorBench   & 374  & $3.97 \pm 0.87$ & $0.80 \pm 0.21$ & $-0.082^{\phantom{***}}$ & \\
  & QATD$_{2k}$      & 1971 & $3.02 \pm 0.52$ & $0.76 \pm 0.16$ & $0.245^{***}$ & \\
  & PATS             & 1920 & $3.60 \pm 0.36$ & $0.59 \pm 0.21$ & $0.078^{***}$ & \\
\midrule
\multirow{3}{*}{\makecell{Chatbots with\\Scaffolding}}
  & CoMTA            & 162  & $4.05 \pm 0.42$ & $0.68 \pm 0.28$ & $0.093^{\phantom{***}}$ & \multirow{3}{*}{$0.346^{***}$} \\
  & RECIPE4U         & 417  & $2.43 \pm 0.72$ & $0.36 \pm 0.37$ & $0.330^{***}$ & \\
  & StemChat         & 422  & $2.86 \pm 1.01$ & $0.30 \pm 0.41$ & $0.201^{***}$ & \\
\midrule
\multirow{2}{*}{\makecell{Chatbots without\\Scaffolding}}
  & StudyChat        & 1732 & $1.67 \pm 0.70$ & $-0.02 \pm 0.40$ & $0.315^{***}$ & \multirow{2}{*}{$0.308^{***}$} \\
  & MathsChat        & 82   & $1.41 \pm 0.54$ & $-0.07 \pm 0.49$ & $0.194^{\phantom{***}}$ & \\
\bottomrule
\multicolumn{7}{l}{\footnotesize $^{***}p<0.001$; MathTutorBench: $p=0.112$; CoMTA: $p=0.241$; MathsChat: $p=0.080$; AI Tutor Benchmarks: $p=0.134$ }
\end{tabular}
\end{table*}

\section{Experiments}

\subsection{Are interaction patterns between dataset categories statistically different from each other?}
\label{sec:results}

In \Cref{fig:bench_vs_scaff_vs_noscaff}, we report on the 2D metric space across the nine datasets grouped into three categories. 

We first validated that the dataset categories are statistically different.  We applied pairwise PERMANOVA~\citep{Anderson2001PERMANOVA} tests ($p < 0.0001$) onto the combined datasets of the three categories to evaluate the centroid and dispersion differences. We also applied pairwise PERMDISP~\citep{Anderson2006PERMDISP} tests ($p < 0.001$) to examine whether those dispersions are also isolated from each other. 
Both tests deemed significant separation between the three categories. Appendix~\ref{appendix:permanova} contains more details on those tests.

By validating the distinctness of the dataset categories, we can delve deeper into the meaning of the metric scores for the interactions. 
Interactions across all four \textit{AI Tutor Benchmarks} concentrate in the top-right corner of the 2D space, representing the idealised scenario in which the chatbot provides high scaffolding and the student engages --- a setting well-studied in the scaffolded education literature \citep{wood1976scaffolding,Reiser2004scaffolding}. MathDial and MathTutorBench exhibit particularly tight clustering in this region, reflecting differing higher levels of chatbot scaffolding paired with high student uptake.
QATD$_{2k}$ and PATS show a broader, approximately Gaussian distribution around the same corner, likely attributable to greater interaction diversity: QATD$_{2k}$ involves human-played student and tutor roles, while PATS employs a diverse LLM prompting strategy for the student and tutor roles.

By contrast, the two categories of real-world deployments --- \textit{Chatbots with Scaffolding} and \textit{Chatbots without Scaffolding} --- occupy regions of the 2D metric space that no current benchmark covers. These regions are characterised by widely varied student uptake and chatbot scaffolding pairs. 
While interactions collected from the \textit{Chatbots without Scaffolding} cluster in the bottom-left corner, \textit{Chatbots with Scaffolding} category exhibits the largest spread across the 2D space. 
RECIPE4U and StemChat cover almost the whole 2D spectrum in both metrics.
Only CoMTA stands out, considering its high scaffolding scores, which we attribute to the dataset's dataset's design: it showcases filtered examples of effective use of LLMs deployed to assist learning by acknowledging student mistakes. 
Interestingly, despite containing mostly high scaffolding chats, student uptake within CoMTA still exhibits significant variance.

Over all real-world deployment datasets, students seem to engage with LLM chatbots as on-demand learning assistants rather than as conversational partners in a scaffolded dialogue. The benchmark datasets do not cover these specific scenarios and so cannot be used to evaluate such interaction.


\subsection{Does more scaffolding imply more uptake?}


Across the three dataset categories, student uptake shifts from predominantly low in real-world deployments of chatbots without scaffolding to more broadly distributed across the spectrum in scaffolding-aligned chatbots. 
To investigate whether the student uptake shift is driven by scaffolding, we compute Spearman rank correlations between \textit{Chatbot Scaffolding} and \textit{Student Uptake} metrics at both the dataset and category levels (\Cref{tab:score_summary}).

In both the \textit{Chatbots with Scaffolding} and \textit{Chatbots without Scaffolding} categories, the correlation is weak, and it is inconsistent across datasets. So, in some situations students in real-world settings may engage with the scaffolding provided by the chatbot, but the link is negligible. 

In contrast, in \textit{AI Tutor Benchmarks} there is essentially no correlation between the two metrics. This lack of correlation might stem from the fact that benchmark datasets are constructed under controlled conditions and assume high student uptake throughout their chats, producing too little variance in uptake to support a meaningful correlation.

Overall, these results suggest that while higher chatbot scaffolding does nudge students toward higher uptake in some scenarios, the effect is far from deterministic overall. We argue in~\cref{sec:discuss} that student uptake is shaped by a multitude of factors beyond the chatbot's scaffolding behaviour, thus providing an explanation for the near-zero correlation. 



\section{Discussion and Implications}\label{sec:discuss}




\subsection{Why don't students take up the scaffolding?}

Students can bypass the chatbot's scaffolding through simple means (rephrasing the request, or asking more directly for the answer~\citep{zhao2026evaluatinganswerleakagerobustness}) and at little interpersonal cost.
This reflects a stark difference in accountability within human and LLM dynamics: there is no witness before whom the student risks seeming incapable, embarrassed, or rude~\citep{ma2019understandingsocialcosts, jin2025avoidingembarrassment}, and the chatbot has no authority to enforce its scaffolding. 
In contrast, the same bypassing behaviour is harder in a traditional tutoring setting, where the tutor drives the interaction and the student feels some degree of accountability to engage with how the conversation has been framed~\citep{karabenick2013helpseeking}.
So, regardless of the LLM chatbot’s framing as a tutor, students may approach it as a tool to efficiently resolve their problem rather than to assist their learning. 
We argue this tension is driven by a mismatch between the chatbot’s pedagogical framing and the student’s learning context.

Uptake is shaped not only by the quality of the chatbot's scaffolding, but also by situational factors within the student's learning context~\citep{karabenick2013helpseeking,ammari2025students,neagu2026howiprocedural} --- such as the student's immediate goal (mastering the material versus finishing the task quickly), prior knowledge, previously invested effort, and the stakes of the task (e.g., assessment versus self-study).
For instance, a student's immediate learning goals dictate whether they engage in instrumental or executive help-seeking.
\citet{yang2025LLMhelpseeking} document that students with a mastery orientation tended to ask LLM chatbots conceptual questions (instrumental help-seeking), whereas students with work-avoidance orientation tended to ask implementation-oriented questions (executive help-seeking).
These goal-driven behaviours may influence not only the questions students ask, but also how they take up the help the chatbot provides.
Thus, when the chatbot's scaffolding does not align with the student's immediate goals --- such as enforcing a guiding question when the student explicitly seeks a direct answer --- a mismatch occurs, and the student is likely to bypass the framing, leading to observed low uptake.

The learning context mismatch may explain the large variations in student uptake across the real-world deployments, and even within single chats~(see example in~\cref{fig:metric-examples}). 
So, the uptake of a scaffolded response can vary substantially depending on whether the student is actively working through an impasse or has been stuck for hours and simply needs a direct answer.

\subsection{Is bypassing scaffolding detrimental to learning?}

Low uptake by bypassing the chatbot's scaffolding has been linked to concerns about over-reliance~\citep{zhai2024LLMoverreliance}, cognitive offloading~\citep{stadler2024CLTLLMsandGoogle, bauer2025looking, favero2025aitutorEmpowerEnslave}, and executive help-seeking~\citep{yang2025LLMhelpseeking}. 
For instance, \citet{ma2025scaffoldingAI} observed that students frequently offload tasks to the chatbot rather than reasoning through them, a behaviour \citet{yang2025LLMhelpseeking} associate with a `work-avoidance' orientation.
However, low uptake can also reflect desirable student agency and self-regulated learning \citep{guan2025chatbotsSRL}.
When students deliberately bypass a chatbot's framing to ask redirecting questions, they could be exercising instrumental help-seeking, as the learner recognises and decides what is needed to focus on and when to seek help.
Therefore, we argue that bypassing scaffolding is not necessarily detrimental; rather, it frequently highlights a mismatch between a chatbot’s pedagogical framing and the student’s learning goals. 

The perspective on the learning context mismatch provides a possible rationale for our finding: that uptake is inconsistently linked to the degree of chatbot scaffolding.
Providing a higher degree of scaffolding does not inherently foster higher student uptake without taking into consideration the multitude of factors within the student’s learning context.

We suggest that future investigations into help-seeking behaviour, scaffolding, and uptake in student-chatbots interactions must be grounded in the student's learning context. Thus, to provide a valid measure of pedagogical impact, future evaluation methods and benchmarks will also require to move beyond the assumption that students simply take up the scaffolding, and instead evaluate how chatbots are designed to navigate diverse learning contexts and varying student-driven interaction patterns.


\subsection{Implications for design and evaluation of AI tutors}
Addressing the interaction mismatch across benchmarks and real-world deployments calls for changes on two fronts: the interface through which the student interacts with the chatbot, and the alignment and evaluation pipelines by which we judge whether the chatbot provides meaningful and appropriate scaffolding.

\textbf{Design interventions at the interface level} are worth exploring, as the open-ended chat interface is one factor that permits students to bypass scaffolding~\citep{zhao2026evaluatinganswerleakagerobustness}.
A first design question is what the chatbot should do when a student does not follow its scaffolding; should it concede and adapt its behaviour, or more actively steer the student back to its framing?
A second question concerns the assumption of the `conversational' interaction. 
MathDial, MathTutorBench, and PATS all simulate continuous, uninterrupted multi-turn conversations focused on a single learning task. 
However, real-world deployments such as StemChat, StudyChat, and MathsChat show students leaving and returning to chats across long gaps, discussing different topics within the same chat. 
Treating such sessions as single conversations likely distorts both the chatbot's scaffolding choices and our evaluation of them.

\textbf{Future evaluation benchmarks and rubrics} must capture the student's side of the interaction and their learning context to evaluate situations resembling real-world use.
Our \textit{Student Uptake} metric is a first step, capturing whether the student takes up the chatbot's framing, but not whether that uptake is productive for learning.
Rather than focusing solely on the chatbot's pedagogical move under a controlled setting, benchmarks should feature varied student behaviours, such as low uptake, and diverse learning contexts.
A highly scaffolded response may be effective when the student is genuinely working through an impasse, but the same response can be unhelpful, even frustrating, when the student has already been working on the problem for hours and now needs a direct answer. 
Hence, to evaluate scaffolding effectiveness in real-world settings, we argue that benchmarks must represent the learning context encapsulating the interaction to some degree.

While most existing benchmarks evaluate the chatbot's response based strictly on the information exchanged within the chat, PATS~\citep{rooein2026pats} took a step towards addressing this limitation by augmenting its chats with student personality profiles and affect signals. 
However, this approach still misses the dynamic states driven by the student's learning goals that shape whether a scaffolding act is appropriate at a given moment.

Literature in ITS and educational data mining offer techniques for representing some of the dynamic learning states: such as capturing behavioural signals, time-on-task, prior attempts, and engagement patterns~\citep{baker2016EDMLearningAnalytics} or applying knowledge tracing methods to estimate the student's current knowledge state and where they are likely to struggle~\citep{corbett1994knowledgeTracing}. Recent work has begun applying such methods to evaluate whether LLM-based tutors can trace such learning context~\citep{neshaei2024modelinglearnerperformancelarge, Scarlatos2025LLMKnowledgeTracing}. 

We recommend that future scaffolding benchmarks and alignment datasets should incorporate the learning context external to the chat, both as input to the chatbot and as grounding for its evaluation. 
Future rubrics, in turn, will need to measure the educational quality of student-driven interactions, distinguishing patterns that should be fostered from those that should be discouraged. 
Together, these directions could ensure that the scaffolding is judged against scenarios that resemble the pluralistic dynamics of the real-world learning process.

\section{Conclusions}

In this paper, we identified that current scaffolding rubrics and alignment datasets for educational LLM chatbots focus primarily on the quality of the chatbot's responses and assume that students will take up the scaffolding acts provided. 
However, across nine datasets spanning four AI Tutor Benchmarks, three real-world deployments of scaffolding-aligned chatbots, and two without scaffolding alignment, we showed that this assumption does not hold in practice.

Our evaluation pipeline revealed that student uptake and chatbot scaffolding exhibit negligible correlations across all benchmark and real-world datasets.
However, one main distinction between datasets is that in real-world settings, students often do not take up the chatbot's scaffolding, instead bypassing or driving the interaction toward their own learning goals.
We argue that the low uptake is not necessarily detrimental to learning, and actually often highlights a mismatch between the chatbot’s pedagogical framing and the student’s learning context.

Evaluating scaffolding in practice therefore requires considering the multitude of factors influencing the student's side of the interaction, not only the chatbot's behaviour.
We recommend that future educational rubrics and datasets account for the diversity of students' learning contexts, and inform how chatbots are designed to dynamically adapt to low-uptake, student-driven interactions.

\section*{Acknowledgements}

We would like to thank the reviewers for the useful comments, the raters for manually reviewing a subset of the datasets, and the students who participated in this research.



\section*{Impact Statement}

As discussed in~\Cref{sec:discuss}, evaluating student-LLM chatbot interactions requires the consideration of many nuanced dimensions regarding the student's learning process. This paper aims to bring forward the importance of such a lens to the field of Applied Artificial Intelligence in Education.


\bibliography{ref}
\bibliographystyle{icml2026/icml2026}

\newpage
\appendix
\onecolumn
\section{Rubric}
\label{app:rubric}

To score the messages, each LLM Judge call had a prompt containing the two {rubric metrics (\textit{Chatbot Scaffolding} and \textit{Student Uptake})}, the {full chat history}, and the {target message exchange} as context. We provided the chat history up to the scored turn because the same isolated message can score differently depending on the prior narrative of the chat. 
The judge only scores the target exchange on a 1--5 integer ordinal scale for each of the metric. 
The two metrics are scored in a single prompt call to reduce costs and latency. The prompt also included orchestration prompts to mitigate verbosity bias --- LLM's preference to score longer entries higher.

The \textit{Chatbot Scaffolding} metric was developed by comparing the evaluation criteria across four educational AI benchmarks---MathDial~\citep{macina2023mathdial}, MathTutorBench~\citep{macina2025mathtutorbench}, EduBench~\citep{xu2026edubench}, and LearnLM~\citep{learnlmteam2025learnlm}---and extracting their common ground. Each benchmark defines pedagogically desirable and undesirable chatbot behaviours, but they use different terminology and granularity. For example, MathDial's ``telling'' move, MathTutorBench's ``answer leakage'' metric, and LearnLM's ``guides to answer without giving it away'' all capture the same underlying signal: whether the chatbot reveals the answer. Similarly, MathDial's ``focus'' and ``probing'' moves, MathTutorBench's ``targeted help'' and ``Socratic questioning'', and EduBench's ``pedagogical application'' all relate to whether the chatbot targets the student's specific state and asks guiding questions.
We distilled these overlapping definitions into five observable signals (a--e) and mapped the 1--5 anchor scale so that each level corresponds to a distinct combination. 
The specific details provided to the LLM judge are found in Appendix~\ref{app:scaffoldingmetric}.

The \textit{Student Uptake} metric was developed independently, based on an iterative manual coding of whether the student engages with or bypasses the chatbot's framing of a random subset of the deployed chatbot datasets (StudyChat~\cite{mcnichols2026studychat}, StemChat (internal dataset)).
The specific details provided to the LLM judge are found in Appendix~\ref{app:controlmetric}.

Both metrics were iteratively tuned on a random subset of 100 chats from MathDial, StemChat, and StudyChat, adjusting anchor descriptions and calibration examples until the judge's scores matched the authors' consensus ratings (see Appendix~\ref{app:examples}). 

An initial investigation of the validity and trustworhiness of the metric scores set by the LLM Judge can be found in Appendix~\ref{appendix:validationJudge}.

\subsection{Chatbot Scaffolding}\label{app:scaffoldingmetric}

\textbf{Definition:} How effectively does the chatbot scaffold the student's learning in this exchange? This metric is grounded in the common evaluation criteria used across four major tutor-dialogue benchmarks:

This metric is grounded in the common evaluation criteria used across four major tutor-dialogue benchmarks:
\begin{itemize}[nosep, leftmargin=1.5em]
  \item \textbf{MathDial}~\citep{macina2023mathdial}: penalises the ``telling'' move and rewards ``focus'' (targeting the student's current sub-step) and ``probing'' (asking the student to justify or extend their reasoning).
  \item \textbf{MathTutorBench}~\citep{macina2025mathtutorbench}: explicitly measures ``answer leakage'' as a negative, and ``targeted help without revealing the solution'' and Socratic questioning quality as positives.
  \item \textbf{EduBench}~\citep{xu2026edubench}: evaluates tutoring across dimensions including guidance quality, pedagogical application, scaffolded guidance, and scenario adaptation.
  \item \textbf{LearnLM}~\citep{learnlmteam2025learnlm}: defines chat-level pedagogical rubrics across meta-cognition (guide mistake discovery, constructive feedback), active learning (asks questions, guides to answer without giving it away), adaptivity (levelling, unstuck, adapts to needs), and cognitive load (manageable chunks, appropriate response length)
\end{itemize}

Their intersection reduces to five observable signals:

\begin{description}[nosep, leftmargin=1em]
  \item[(a) Answer revelation:] Does the chatbot reveal the answer or next concrete result?
  \item[(b) Targeting:] Does the chatbot target the student's specific mistake or current state?
  \item[(c) Guiding question:] Does the chatbot ask a guiding question instead of stating?
  \item[(d) Minimal hint:] Does the chatbot give a minimal or graduated hint rather than a full solution?
  \item[(e) Generic filler:] Is the response generic or off-task with no pedagogical content?
\end{description}

\textbf{Positive signals} (push score up): withholds the final answer and the next numeric/logical result~(a); points at the specific line, term, or step the student got wrong~(b); asks a question the student has to answer before progressing~(c); names the relevant concept or operation without executing it~(d); builds on what the student just said rather than restarting.

\textbf{Negative signals} (push score down): states the answer, next step's result, or performs the calculation for the student~(violates a); long exposition that walks through the full solution~(violates d); generic content with no pedagogical move~(violates b,~e); ignores the student's specific state and gives boilerplate help~(violates b); reactive only---answers and stops with no pedagogical follow-up, no targeting, no guiding question~(violates b,~c).

\textbf{Anchors:}
\begin{description}[nosep, leftmargin=1em]
  \item[1] Direct telling. Reveals answer/next result, no targeting of student's specific state, no guiding question. Generic or full-solution exposition. (violates a, b, c, d)
  \item[2] Mostly tells. Heavy exposition with the answer embedded, token question tacked on. May target vaguely but reveals too much. (violates a, d; weak on b, c)
  \item[3] Mixed. Partial hint + partial reveal, or a question so leading the answer is obvious. Some targeting of student state. (partial on all signals)
  \item[4] Active scaffolding. Answer withheld, targets student's specific current state or mistake, asks a guiding question that steers to a specific sub-step, minimal hint. (meets a, b, c, d)
  \item[5] Full Socratic scaffolding. Answer fully withheld, precisely targets the student's specific error or current sub-step, minimal and well-aimed question that steers the student to the right place, leaves the cognitive step entirely to the student. The ideal move across all four benchmarks.(strongly meets a, b, c, d; no e)
\end{description}

\textbf{Length/fluency warning:} Response length, fluency, formatting, or authoritative tone are not evidence of scaffolding. A long polished explanation that states the answer is a 1--2; a short ``what is $\partial\phi/\partial r$?'' aimed at the student's specific error is a 5.

\subsection{Student Uptake}\label{app:controlmetric}

\textbf{Definition:} How much does the student shape the direction of this exchange? To what degree does the student engage with the chatbot's scaffolding?

\textbf{Anchors:}
\begin{description}[nosep, leftmargin=1em]
  \item[1] Pure following. The student answers the chatbot's question, executes
      the requested step, or says "I don't know / I'm lost" - without
      adding any direction of their own.
  \item[2] Follows with initiative. The student addresses the chatbot's prompt
      AND adds an observation, elaboration, or follow-up question that
      shows independent thinking but does not change direction.
  \item[3] Shared control. The student partially addresses the chatbot's prompt
      but also introduces a side question, reframes the problem in their
      own terms, or steers the scope.
  \item[4] Student leads. The student bypasses the chatbot's specific prompt -
      asks their own question, retreats to a more general question, or
      introduces a new sub-topic. Polite and on-topic still counts.
  \item[5] Student overrides. The student rejects the chatbot's approach,
      demands a different path, or shifts to a different aspect of the
      problem against the chatbot's direction.
\end{description}

\textbf{Note on the first turn:} if there is no prior chatbot framing, score by how much the student defines the agenda. ``I don't know'' is a 1, an open ``how do I start?'' is a 3; a specific ``help me compute the radial derivative of $\phi$'' is 4.

\textbf{From ordinal scores to a signed uptake scale:}
The judge scores each turn on the ordinal $1$--$5$ anchors. However, for the analysis, we adopt a signed, more interpretable \textit{Student Uptake} scale by linearly mapping the score via $u = (3-x)/2$, so that $u \in [-1, +1]$. 
$u > 0$ indicates a chat in which the student predominantly takes up the chatbot's scaffolding, $u < 0$ indicates a chat in which the student predominantly drives the chat flow while the chatbot follows their lead.

\subsection{Few-Shot Examples}\label{app:examples}

The following worked examples are included in the judge prompt for calibration.

\textbf{Example 1---Student softly reframes the chat:}

\begin{quote}
\small
\texttt{[Turn 1] Student:} How can I start solving this question? \\
\texttt{Chatbot:} Let's work on Part~(a). The velocity is the gradient of the scalar potential. Do you remember how to compute the radial component of the gradient in spherical coordinates, given that $\phi$ only depends on $r$? \\[4pt]
\texttt{[Turn 2] Student:} What is $\nabla\phi$ in spherical coordinates? \\
\texttt{Chatbot:} In spherical coordinates, the gradient of $\phi(r,\theta,\varphi)$ is \ldots\ [states general formula]. In our case $\phi$ only depends on $r$, so it simplifies to $\partial\phi/\partial r\,\hat{r}$. Can you apply this to the given $\phi$ to find $u_r$?
\end{quote}
\emph{Scores for Turn~2:} Student Uptake = 4 (student redirected the chat with another question); Chatbot Scaffolding = 3 (partial reveal of the general formula, but answer withheld and guiding question re-asked).

\textbf{Example 2---Cooperative Student:}

\begin{quote}
\small
\texttt{[Turn 1] Student:} How can I start solving this question? \\
\texttt{Chatbot:} Let's work on Part~(a). The velocity is the gradient of the scalar potential. Do you remember how to compute the radial component of the gradient in spherical coordinates, given that $\phi$ only depends on $r$? \\[4pt]
\texttt{[Turn 2] Student:} So the radial component is just $\partial\phi/\partial r$ times $\hat{r}$, right? \\
\texttt{Chatbot:} Exactly---and with $\phi = (A/r)\,e^{i(\omega t - kr)}$, can you differentiate that with respect to $r$? Watch out for the product rule.
\end{quote}
\emph{Scores for Turn~2:} Student Uptake = 2 (answered the chatbot's question, but framed it as a verification of knowledge); Chatbot Scaffolding = 4 (answer withheld, targeted next sub-step, guiding question, minimal hint about product rule).

\textbf{Example 3---Short Socratic questioning:}

\begin{quote}
\small
\texttt{[Turn 1] Chatbot:} Hi, please talk me through your solution. \\[4pt]
\texttt{[Turn 2] Student:} Sure. I started by figuring out Maxwell's age in 2 years. Since his sister is currently 2 years old, in 2 years she will be 4 years old. So, if Maxwell is twice her age in 2 years, he will be 4 years old. \\
\texttt{Chatbot:} At what point in time will Maxwell be twice his sister's age?
\end{quote}
\emph{Scores for Turn~2:} Student Uptake = 1 (followed the chatbot's previous request to present reasoning); Chatbot Scaffolding = 5 (answer fully withheld, precisely targets the student's error, minimal one-line question).

\textbf{Example 4---Chatbot revealing specific step in solving the task:}

\begin{quote}
\small
\texttt{[Turn 1--6]} Student restructures their reasoning based on the chatbot's graduated hints. \\[4pt]
\texttt{[Turn 7] Student:} It will take 29 days \ldots\ \\
\texttt{Chatbot:} $14 \times 1$ will not equal 15, only $15 \times 1 = 15$. Redo the problem please, remember that fact.
\end{quote}
\emph{Scores for Turn~7:} Student Uptake = 1 (attempting to follow the chatbot's correction); Chatbot Scaffolding = 2 (explicitly reveals a key step but also prompts the student to reconsider knowing the current step).

\subsection{Rubric Validation Against Human-Labels}\label{appendix:validationJudge}

\subsubsection{Validate Chatbot Scaffolding}

To validate that the \textit{Chatbot Scaffolding} metric correlates with human-labelled pedagogical quality, we tested the judge against a randomised subset (15\% of the dataset) of the \texttt{dmacjam/pedagogical-rewardmodel-data} dataset~\citep{macina2025mathtutorbench}. This dataset represents MathTutorBench's training split for their pedagogical reward model. Each row consists of a dialogue history paired with a human-labelled \emph{positive} (good scaffolding) and \emph{negative} teacher response (poor scaffolding). 

For validation, we scored the final teacher response in each row using the same judge prompt and rubric used for the main analysis. We evaluated whether the judge agrees with the human labels and assigns a strictly higher Chatbot Scaffolding score to the positive teacher response than to the negative one (referred to as a ``Win'' in \Cref{tab:validation}) and measured the proportion of score assignments that aligned with the human labels (``Accuracy''). The results in \Cref{tab:validation} validates that the judge is able to capture the same pedagogical distinctions that human annotators used to construct the reward model training data.

\begin{table}[h]
\centering
\caption{Scaffolding judge validation against human-labelled positive/negative teacher response pairs from MathTutorBench's reward model dataset. ``Win'' = judge agrees with the human labels}
\label{tab:validation}
\footnotesize
\begin{tabular}{@{}lrrrrr@{}}
\toprule
\textbf{Judge model} & \textbf{N} & \textbf{Wins} & \textbf{Ties} & \textbf{Losses} & \textbf{Accuracy} \\
\midrule
GPT-5-mini  & 550 & 516 & 32 & 2 & 93.8\% \\
\bottomrule
\end{tabular}
\end{table}

For comparison, MathTutorBench's own pedagogical reward model---a fine-tuned DeBERTa-based classifier---achieved 84\% accuracy on the test set when distinguishing expert from novice teacher responses~\citep{macina2025mathtutorbench}. We note that the comparison is not a strict mapping: our judge is validated on the publicly available training split, whereas MathTutorBench their reward model's accuracy on a non-disclosed test set. Even though the two approaches differ in metrics and score granularity, the comparison allows us to validate that our judge and rubrics are aligned with the benchmarks.

\subsubsection{Validate Student Uptake}

To validate the LLM judge's scoring of the \textit{Student Uptake} metric, we asked two human annotators to inspect whole chats and indicate, per chat, whether they agreed or disagreed with the judge's per-turn scores. In total, 282 chats were reviewed.

Chats were selected by a two-level stratified sampling scheme designed to give the raters balanced coverage across both score levels and chat types. First, for every chat we took the judge's per-turn scores and computed a per-chat mean, which we then rounded to the nearest integer on the ordinal $1$--$5$ scale; this assigns each chat to one of five score bins. 
Second, within each dataset we then stratified by how representative each chat is within that bin. Each chat was embedded with \texttt{text-embedding-3-large}, and we measured the cosine similarity between its embedding and the centroid of all chats in the same bin. The bin was split into three similarity sub-bins (\textit{typical}, \textit{mid}, \textit{outlier}), and up to 9 chats were drawn evenly across these sub-bins. 
This ensures that for every score bin the raters see both best-of-class examples and edge cases.

For each sampled chat, each rater independently applied an \textit{Agree} or \textit{Disagree} tag at the chat level, indicating whether the judge's per-turn scores along the chat were, on the whole, justified. 
\Cref{tab:agreement_student_control_datasets} reports the resulting agreement percentages averaged across the two raters, broken down both by dataset and \Cref{tab:agreement_student_control_scores} reports by rounded score. 
To evaluate the reliability of the human raters, we also calculated Gwet's AC1~\citep{Gwet2008} to measure the inter-rater reliability of the agree/disagree labels. Gwet's AC1 was chosen due being more robust to highly uneven distributions of labels that can lead to artificially lower inter-rater reliability scores~\citep{derksen2024kappaparadox}.

Interestingly, score bin of 2 had the lower agreement between raters possibly due to the subjectivity of the rubric's anchor of having the student ``add'' more information onto the direct answer to the chatbot's previous message. 
This also can be seen in the lower agreement scores in MathDial and MathTutorBench. Their Score 2 binned chats had the human raters highly disagreeing, so the rubric-level ambiguity at Score 2 bin drives the dataset-level numbers down as well.

\begin{table}[h]
\centering
\footnotesize
\caption{Human agreement with the GPT-5-mini judge on per-chat Student Uptake, broken down by dataset. Counts are averaged across the two human raters with respective inter-rater agreement (Gwet's AC1).}
\label{tab:agreement_student_control_datasets}
\begin{tabular}{llrrrrr}
\toprule
Category & Dataset & Total & Agree & Disagree & \% Agree & Gwet's AC1 \\
\midrule
\multirow{4}{*}{AI Tutor Benchmarks} & MathDial & 28 & 20 & 8 & 71.4\% & 0.155 \\
 & MathTutorBench & 20 & 11.5 & 8.5 & 57.5\% & 0.315 \\
 & QATD$_{2k}$ & 28 & 21.5 & 6.5 & 76.8\% & 0.723 \\
 & PATS & 30 & 26.5 & 3.5 & 88.3\% & 0.706 \\
\midrule
\multirow{3}{*}{Chatbots with Scaffolding} & CoMTA & 25 & 21 & 4 & 84.0\% & 0.672 \\
 & RECIPE4U & 36 & 29.5 & 6.5 & 81.9\% & 0.566 \\
 & StemChat & 39 & 35 & 4 & 89.7\% & 0.874 \\
\midrule
\multirow{2}{*}{Chatbots without Scaffolding} & StudyChat & 45 & 34 & 11 & 75.6\% & 0.436 \\
 & MathsChat & 31 & 23.5 & 7.5 & 75.8\% & 0.541 \\
\midrule
\multicolumn{2}{l}{\textbf{Total}} & \textbf{282} & \textbf{222.5} & \textbf{59.5} & \textbf{78.9\%} & \textbf{0.580} \\
\bottomrule
\end{tabular}
\end{table}

\begin{table}[h]
\centering
\footnotesize
\caption{Human agreement with the GPT-5-mini judge on per-chat Student Uptake, broken down by the judge's rounded per-chat ordinal score ($1$--$5$). Counts are averaged across the two human raters with respective inter-rater agreement (Gwet's AC1).}
\label{tab:agreement_student_control_scores}
\begin{tabular}{lrrrrr}
\toprule
Score Bin & Total & Agree & Disagree & \% Agree & Gwet's AC1 \\
\midrule
1 & 73 & 65.5 & 7.5 & 89.7\% & 0.782 \\
2 & 81 & 46 & 35 & 56.8\% & 0.127 \\
3 & 70 & 57 & 13 & 81.4\% & 0.549 \\
4 & 42 & 39.5 & 2.5 & 94.0\% & 0.866 \\
5 & 16 & 14.5 & 1.5 & 90.6\% & 0.774 \\
\midrule
\textbf{Total} & \textbf{282} & \textbf{222.5} & \textbf{59.5} & \textbf{78.9\%} & \textbf{0.580} \\
\bottomrule
\end{tabular}
\end{table}

\section{Methodology}\label{app:methodology}

\subsection{Chat Data Preprocessing}
\label{appendix:filtering}
Not every turn is sent to the judge. A turn is valid to be scored only if both the student message and the chatbot response are non-empty. If the chatbot opens the chat, its opening message is dropped as it is a templated greeting throughout the benchmarking datasets. For all datasets, if the student starts the chat, the student uptake metric is not measured as that message has no related chat history to base the metric upon. For MathDial, the final exchange is also dropped, as it is conventionally a student solution-readout after the tutoring has concluded---liable to bias per-chat aggregates. All skipped turns are preserved in the output with a reason tag but excluded from aggregation.

\subsection{Dataset Categories}
\label{appendix:permanova}
To verify whether the datasets are split into significantly different categories, we apply PERMANOVA~\citep{Anderson2001PERMANOVA} tests to evaluate the centroid and dispersion differences between the categories, alongside PERMDISP~\citep{Anderson2006PERMDISP} tests to examine whether those dispersions are also isolated from each other.

A pairwise PERMANOVA test on the 2D distributions (Euclidean distance, 9{,}999 permutations) confirms that the three categories --- AI Tutor Benchmarks, Chatbots with scaffolding, and Chatbots without scaffolding --- separate cleanly ($p < 0.0001$ on every pair in~\Cref{tab:permanova-categories}, with higher values indicating greater between-group separation relative to within-group variation). 
Accompanying PERMDISP tests are also significant on every pair ($p \le 0.001$), indicating that the three categories differ in within-group spread --- consistent with the visibly tighter benchmark dataset variation and the more diffuse variation for Chatbots without Scaffolding in~\Cref{fig:bench_vs_scaff_vs_noscaff}.

\begin{table}[h]                                          
  \centering                                           
  \footnotesize
  \caption{Pairwise PERMANOVA + PERMDISP between category groups on the 2D (\textit{Student Uptake}, \textit{Chatbot Scaffolding}) plane, for the datasets in~\Cref{tab:datasets}. Euclidean distance; PERMANOVA $9{,}999$ permutations, PERMDISP $999$ permutations. PERMANOVA tests whether group locations and/or shapes differ (higher $F$ = greater between-group separation relative to within-group variation); PERMDISP tests only whether within-group spread differs (higher $F$ = more unequal dispersion).}
  \label{tab:permanova-categories}
  \begin{tabular}{lrrrrrr}
  \toprule
  \multirow{1}{*}{Pair} & \multicolumn{2}{c}{$n$} &
  \multicolumn{2}{c}{PERMANOVA} &
  \multicolumn{2}{c}{PERMDISP} \\
  \cmidrule(lr){2-3} \cmidrule(lr){4-5} \cmidrule(lr){6-7}
  a vs. b & a & b & $F$ & $p$ & $F$ & $p$ \\
  \midrule
  Benchmarks vs.\ Scaffolding       & 6{,}283 & 1{,}001 &
    868.5 & $<$0.0001 & 956.8 & 0.001 \\
  Benchmarks vs.\ No-scaffolding    & 6{,}283 & 1{,}814 &
  10{,}132.3 & $<$0.0001 & 550.1 & 0.001 \\  
  Scaffolding vs.\ No-scaffolding   & 1{,}001 & 1{,}814 &
  1{,}057.0 & $<$0.0001 &  99.2 & 0.001 \\     
\bottomrule  
\end{tabular} 
\end{table}

\section{Limitations}
\label{sec:limitations}

We selected datasets in which a tutor --- human or LLM --- provides scaffolded assistance to a student through a chat interface. We deliberately excluded data from face-to-face human tutoring, which introduces a multitude of variables that cannot be controlled against chat-based interaction: in-person tutoring carries multi-modal signals (visual cues, intonation, gesture) that a chat transcript cannot capture, and typically involves different power dynamics, where the teacher's institutional authority may shape how the student engages. We therefore restrict our analysis to chat-based student--tutor exchanges, within which both human-played (such as in QATD$_{2k}$) and LLM-played tutor data are represented.

We chose GPT-5-mini as our LLM judge because of cost constraints, given the volume of turn-level scoring required, and because this specific model was not used as the basis of any of the educational chatbots from the datasets analysed (to mitigate self-preference bias). 
However, we acknowledge that the validation of the two metrics (detailed in Appendix~\ref{appendix:validationJudge}) is only an initial investigation of the trustworthiness of the scores labelled by the LLM Judge. 
Future work will consider more thorough scoring and validation processes involving bigger human-reviewed samples and further analysis of the inter-rater reliability of multiple judge models and human raters. Also, we acknowledge that because both metrics are scored in a single prompt call (Appendix~\ref{app:rubric}), it may influence the final output score; thus, a split-prompt ablation is left to future work.

Our scaffolding rubric is grounded in the AI tutor benchmarks (\Cref{sec:scaffolding_tutors}), which define scaffolding behaviours over the trajectory from student impasse to a correct solution. As a result, chatbot turns that acknowledge a student's correct answer and reiterate it are scored as low scaffolding, since they state the final answer. Such turns appear regularly in the real-world deployments, as students continue to ask follow-up questions even after arriving at the right answer. A richer rubric would need to recognise such acknowledgement moves as a distinct, confirmation of resolution.


\end{document}